%% file: main.tex
\documentclass{article}
\usepackage[ruled,vlined,linesnumbered]{algorithm2e}
\usepackage{amsmath}
\usepackage{amssymb}
\usepackage{bm}
\usepackage{booktabs}
\usepackage{graphicx}
\usepackage{tikz}
\usepackage{xcolor}

\title{Understanding Gradient Boosting Classifier: Training, Prediction, and the Role of $\gamma_j$}
\author{Hung-Hsuan Chen}
\date{Department of Computer Science and Information Engineering\\
National Central University\\
\texttt{hhchen1105@acm.org}
}

\begin{document}

\maketitle

\input{contents/abs}
\input{contents/intro}
\input{contents/gbc-algo}
\input{contents/gamma-detail}
\input{contents/summary}

\bibliographystyle{apalike}
\bibliography{ref}

\newpage

\appendix

\input{contents/appendix}

\end{document}

%% file: contents/abs.tex
\begin{abstract}

The Gradient Boosting Classifier (GBC) is a widely used machine learning algorithm for binary classification, which builds decision trees iteratively to minimize prediction errors. This document explains the GBC's training and prediction processes, focusing on the computation of terminal node values $\gamma_j$, which are crucial to optimizing the logistic loss function. We derive $\gamma_j$ through a Taylor series approximation and provide a step-by-step pseudocode for the algorithm's implementation. The guide explains the theory of GBC and its practical application, demonstrating its effectiveness in binary classification tasks. We provide a step-by-step example in the appendix to help readers understand.

\end{abstract}

%% file: contents/intro.tex
\section{Introduction}

The gradient boosting machine (GBM)~\cite{friedman2001greedy, hastie2009elements} is a robust predictive model for tabular datasets. GBM constructs weak learners (trees) sequentially; each tree predicts the residual of the earlier predictions.

When applying GBM to regression tasks, the training process of each tree is very similar to that of a standard decision tree. However, when applying GBM on binary classification tasks (we call the Gradient Boosting Classifier method, or GBC, below), the prediction of a terminal node $j$ of a tree $T_m$ is not the average residual of the training instances that fall at node $j$. Instead, the prediction of the terminal node $j$ in the tree $T_m$ is given by a mysterious equation below.

\begin{equation} \label{eq:gamma-j}
    \gamma_{j} = \frac{\sum_{i\in \Omega_{m,j}} r_m^{(i)}}{\sum_{i\in \Omega_{m,j}} p_{m-1}^{(i)} (1-p_{m-1}^{(i)})},
\end{equation}
where $p_{m-1}^{(i)}$ is the estimated probability that the $i$th instance's target value is 1 in the previous iteration, $r_m^{(i)} = y^{(i)} - p_{m-1}^{(i)}$ is the residual of instance $i$ that needs to be fit/predict in iteration $m$, and $\Omega_{m,j}$ is the set of instance indices located in the terminal node $j$ for tree $T_m$. 

In the next section, we introduce the GBC training and prediction procedure, followed by a clear explanation of how $\gamma_j$ is derived in Section~\ref{sec:gamma-details}. We give a step-by-step example in the appendix in Section~\ref{app:example}.

%% file: contents/gbc-algo.tex
\section{GBC training and prediction}

\begin{algorithm}[tbh]
\SetKwInOut{Input}{input}\SetKwInOut{Output}{output}
\Input{Input features $\bm{x}^{(1:n)} \in \mathbb{R}^{n \times d}$ and targets $y^{(1:n)} \in \{0, 1\}^n$}
\Input{Number of Trees: $M$}
\Input{Learning rate: $\eta$}
\Output {Fitted trees: $T_{1:M}$}

\tcp{Initialization}
\For{$i\gets 1$ \KwTo $n$}{
    $F_0^{(i)} \gets 0$\;
    $p_0^{(i)} \gets \sigma\left(F_0^{(i)}\right) = 0.5$\;
}

\tcp{Training for $M$ iterations}
\For{$m \gets 1$ \KwTo $M$} {
    \For {$i\gets 1$ \KwTo $n$}{
        Compute residual: $r_m^{{(i)}} \gets y^{(i)} - p_{m-1}^{(i)}$\;
    }

    Train a regression tree $T_m$ to fit the dataset: $\{(\bm{x}^{(i)}, r_m^{(i)})\}_{i=1,\ldots,n}$\;

    \tcp{Update terminal node values}
    Let $\Omega_{m,j}$ be the set of instance indices in terminal node $j$ of $T_m$\;
    \For {each terminal node $j \in T_m$}{
        Set the output of node $j$ in $T_m$: $\gamma_{j} \gets \frac{\sum_{k\in \Omega_{m,j}} r_m^{(k)}}{\sum_{k\in \Omega_{m,j}} p_{m-1}^{(k)} (1-p_{m-1}^{(k)})}$\;
        \For{$k \in \Omega_{m,j}$}{
            $F_m^{(k)} \gets F_{m-1}^{(k)} + \eta \gamma_{j}$\;
            $p_m^{(k)} \gets \sigma\left(F_m^{(k)}\right)$\;
        }
    }
}
\caption{Gradient Boosting Classifier Training}
\label{alg:gbc-train}
\end{algorithm}

\subsection{Training process}

The training of a Gradient Boosting Classifier begins with an initial model, which is typically a constant prediction for all instances. In binary classification, the model is initialized with a raw predicted logarithm of odds (called log odds below) $F_0^{(i)}$ set to 0 for each instance $i$, which corresponds to an initial predicted probability of 0.5 for all instances (since $p_0^{(i)}=\sigma(F_0^{(i)})$, where $\sigma$ is the logistic function). This represents a neutral starting point since the model does not have prior information about the data.

The algorithm iteratively proceeds over $M$ boosting iterations. At each iteration $m$, the model computes the residuals for each training instance. The residuals represent how far the current predicted probability is from the true label $y^{(i)}$. Specifically, the residual $r_m^{(i)}$ is computed as the difference between the true label $y^{(i)}$ and the predicted probability $p_{m-1}^{(i)}$ from the previous iteration. These residuals are used to train the decision tree $T_m$, where the tree is fitted to predict the residuals, thus focusing the model on correcting the errors made by the previous trees. The tree training strategy could be based on metrics such as information gain, Gini-index, or mean squared error, guiding the tree to select the optimal feature and threshold for splitting~\cite{han2012mining}.

Once the tree $T_m$ is trained, the algorithm computes the optimal values for the terminal nodes. For each terminal node $j$, the output value $\gamma_j$ is calculated by minimizing the logistic loss for the instances that fall into that node. This is done using Equation~\ref{eq:gamma-j}; the reasons will be explained in Section~\ref{sec:gamma-details}.

Next, for each instance $k$ that falls into terminal node $j$, the model prediction is updated by adding the scaled node value $\gamma_j$ to the previous prediction $F_{m-1}^{(k)}$, with a scaling factor given by the learning rate $\eta$. This update is applied as follows:

\begin{equation} \label{eq:logloss-update}
    F_m^{(k)} \gets F_{m-1}^{(k)} + \eta \gamma_{j}    
\end{equation}

The predicted probability of instance $k$ is then updated by applying the logistic function to the new raw score $F_m^{(k)}$:

\begin{equation}
    p_m^{(k)} = \sigma(F_m^{(k)}).
\end{equation}

The algorithm builds trees over $M$ iterations, incrementally refining the model's predictions by reducing residual errors from previous iterations. At the end of the training process, we have an ensemble of $M$ decision trees $T_{1:M}$, each contributing to the final prediction.

Algorithm~\ref{alg:gbc-train} shows the pseudocode for training.

\subsection{Prediction process}

\begin{algorithm}[tbh]
\SetKwInOut{Input}{input}\SetKwInOut{Output}{output}
\Input{Test instances $\bm{x}^{(1:t)} \in \mathbb{R}^{t \times d}$}
\Input{Learning rate: $\eta$}
\Input{Trained trees: $T_{1:M}$}
\Output{Predicted probabilities $p^{(1:t)}$ for class 1}
\For{$i\gets 1$ \KwTo $t$}{
    $F^{(i)} \gets 0$\;
    \For{$m\gets 1$ \KwTo $M$}{
        $F^{(i)} \gets F^{(i)} + \eta T_m(\bm{x}^{(i)})$\tcp*{$T_m(\bm{x}^{(i)})$ is $T_m$'s prediction on $\bm{x}^{(i)}$}

    }
    $p^{(i)} \gets \sigma\left(F^{(i)}\right) = \frac{1}{1 + e^{-F^{(i)}}}$\;
}
\caption{Gradient Boosting Classifier Prediction}
\label{alg:gbc-pred}
\end{algorithm}

Once the Gradient Boosting Classifier is trained, GBC predicts the labels of new test instances as follows: The prediction process starts by initializing the raw prediction $F^{(i)}$ for each test instance $i$ to 0, just as in the training phase. The algorithm then iteratively applies each of the $M$ trees to the test instance, adding the scaled output of each tree to the current raw score.

Specifically, for each tree $T_m$, the prediction for a test instance $\bm{x}^{(i)}$ is obtained by evaluating the decision tree. The output value, $T_m(\bm{x}^{(i)})$, is then scaled by the learning rate $\eta$ and added to the current raw score $F^{(i)}$ as:

\begin{equation}
    F^{(i)} = F^{(i)} + \eta T_m(\bm{x}^{(i)})
\end{equation}

This process is repeated for each tree from $T_1$ to $T_M$, accumulating the contributions of all trees to the final raw prediction $F_i$.

Once all trees have been evaluated, the final predicted probability for the test instance $\bm{x}^{(i)}$ is computed by applying the logistic function to the raw score $F^{(i)}$:

\begin{equation}
    p^{(i)} = \sigma(F^{(i)}) = \frac{1}{1+e^{-F^{(i)}}}
\end{equation}

The output is a probability score for each test instance, indicating the confidence of the model in predicting class 1. This probability can be thresholded (e.g., using a threshold of 0.5) to produce binary class predictions.

Much like the training process, this prediction process is designed to leverage the additive nature of Gradient Boosting, where the final prediction is built up from the contributions of many weak learners.

The pseudocode is shown in Algorithm~\ref{alg:gbc-pred}.

%% file: contents/gamma-detail.tex
\section{Where does $\gamma_{j}$ come from?} \label{sec:gamma-details}

Assume that GBC has been trained for $m-1$ iterations, so $F_{m-1}(\bm{x}^{(i)})$ is fixed. GBC attempts to train a new weak learner $T_m$ that has $J$ terminal node predictions $\gamma_{1}, \ldots, \gamma_{J}$. The new predicted logarithm of odds $F_m(\bm{x}^{(i)})$ is given by

\begin{equation}
    F_m(\bm{x}^{(i)}) = F_{m-1}(\bm{x}^{(i)}) + \gamma_{j},
\end{equation}
where $j$ is the index of the terminal node that $\bm{x}^{(i)}$ falls in.

The predicted probability is

\begin{equation} \label{eq:prob-of-true}
    p_m^{(i)} = \sigma\left(F_m(\bm{x}^{(i)})\right) = \frac{1}{1+e^{-F_m(\bm{x}^{(i)})}} = \frac{1}{1+e^{-\left(F_{m-1}(\bm{x}^{(i)}) + \gamma_j\right)}}
\end{equation}

Let $\bm{y}^{\Omega_{m,j}}$ and $\bm{p}_m^{\Omega_{m,j}}$ denote the ground truth labels and predicted probabilities after iteration $m$ for all the nodes that fall in the terminal node $j$ of the tree $T_m$, we express the cross entropy loss for these instances as a function of $\gamma_j$:

\begin{equation} \label{eq:leave-loss}
\begin{split}
    f(\gamma_j) &= \mathcal{L}(\bm{y}^{\Omega_{m,j}}, \bm{p}_m^{\Omega_{m,j}}) = -\sum_{\forall k \in \Omega_{m,j}} \left(y^{(k)} \log p_m^{(k)} + (1-y^{(k)}) \log (1-p_m^{(k)})\right) \\
    & = \sum_{\forall k \in \Omega_{m,j}} \left(-y^{(k)} \log p_m^{(k)} - \log (1-p_m^{(k)}) + y^{(k)} \log(1-p_m^{(k)})\right) \\
    & = \sum_{\forall k \in \Omega_{m,j}} \left(-y^{(k)} \log \frac{p_m^{(k)}}{1-p_m^{(k)}} - \log (1-p_m^{(k)})\right) \\
    & = \sum_{\forall k \in \Omega_{m,j}} \left( -y^{(k)} \log\left(\frac{1}{e^{-F_{m}(\bm{x}^{(i)})}}\right) - \log\left(1 - \frac{1}{1+e^{-F_{m}(\bm{x}^{(k)})}}\right) \right) \\
    & = \sum_{\forall k \in \Omega_{m,j}} \left( -y^{(k)} F_m(\bm{x}^{(i)})  - \log\left(\frac{e^{-F_{m}(\bm{x}^{(k)})}}{1+e^{-F_{m}(\bm{x}^{(k)})}}\right) \right) \\
    & = \sum_{\forall k \in \Omega_{m,j}} \left( -y^{(k)} F_m(\bm{x}^{(i)}) - \log\left(\frac{1}{1+e^{F_{m}(\bm{x}^{(k)})}}\right) \right) \\
    & = \sum_{\forall k \in \Omega_{m,j}} \left( -y^{(k)} F_m(\bm{x}^{(i)}) + \log\left(1+e^{F_{m}(\bm{x}^{(k)})}\right) \right) \\
    & = \sum_{\forall k \in \Omega_{m,j}} \left( -y^{(k)} \left(F_{m-1}(\bm{x}^{(i)}) + \gamma_j\right)+ \log\left(1+e^{\left(F_{m-1}(\bm{x}^{(k)}) + \gamma_j\right)}\right) \right)
\end{split}
\end{equation}

We look for the $\gamma_{j}$ value to minimize the cross-entropy loss.

\subsection{Trial 1: set the derivative to zero and solve the equation}

The loss function can be considered a function of $\gamma_{j}$ since $\gamma_{j}$ is the only variable in the function. We take the derivative of the loss with respect to $\gamma_{j}$ and set it to zero:

\begin{equation*} 
\begin{split}
\frac{f(\gamma_j)}{\partial \gamma_j} &= \frac{\partial \mathcal{L}(\bm{y}^{\Omega_{m,j}}, \bm{p}_m^{\Omega_{m,j}})}{\partial \gamma_{j}} = \sum_{\forall k \in \Omega_{m,j}}\left( -y^{(k)} + \frac{e^{F_{m-1}(\bm{x}^{(k)}) + \gamma_{j}}}{1+e^{F_{m-1}(\bm{x}^{(k)}) + \gamma_{j}}}\right) \\
&= \sum_{\forall k \in \Omega_{m,j}} \left( -y^{(k)} \right) + \sum_{\forall k \in \Omega_{m,j}} \left( \frac{1}{1+e^{-(F_{m-1}(\bm{x}^{(k)}) + \gamma_{j})}} \right) := 0 \end{split}
\end{equation*}

\begin{equation} \label{eq:derive-set-to-zero}
\Rightarrow \sum_{\forall k \in \Omega_{m,j}} \left( \frac{1}{1+e^{-(F_{m-1}(\bm{x}^{(k)}) + \gamma_{j})}} \right) = \sum_{\forall k \in \Omega_{m,j}} y^{(k)}
\end{equation}

Equation~\ref{eq:derive-set-to-zero} involves $\gamma_j$ within a nonlinear logistic function, making it challenging to isolate $\gamma_j$ and solve explicitly. This motivates the need for an approximation technique, such as a second-order Taylor expansion.

\subsection{Trial 2: Approximate the loss function by Taylor's series}

Taylor's expansion approximates a function $f(x)$ based on a fixed $x_0$ as follows~\cite{canuto2015mathematical}.

\begin{equation} \label{eq:taylor-orig}
    f(x) \approx f(x_0) + (x-x_0) f'(x_0) + \frac{1}{2} (x-x_0)^2 f''(x_0)
\end{equation}

We approximate Equation~\ref{eq:leave-loss}, $f(\gamma_j)$, at $x_0=0$ based on the Taylor expansion:

\begin{equation} \label{eq:taylor-loss}
\begin{split}
f(\gamma_j) & \approx f(0) + (\gamma_j-0)\frac{\partial \mathcal{L}}{\partial \gamma_j}\bigg|_{\gamma_j=0} + \frac{1}{2} (\gamma_j-0)^2 \frac{\partial^2 \mathcal{L}}{\partial \gamma_j^2} \bigg|_{\gamma_j = 0}\\
&= f(0) + \gamma_j \sum_{\forall k \in \Omega_{m,j}} \left(-y^{(k)} + \frac{e^{F_{m-1}(\bm{x}^{(k)})}}{1+e^{F_{m-1}(\bm{x}^{(k)})}}\right) \\
&+ \frac{\gamma_j^2}{2} \sum_{\forall k \in \Omega_{m,j}} \left(\sigma(F_{m-1}(\bm{x}^{(k)}))(1-\sigma(F_{m-1}(\bm{x}^{(k)})))\right) \\
&= f(0) + \gamma_j \sum_{\forall k \in \Omega_{m,j}} \left(-y^{(k)} + p_{m-1}^{(k)}\right) + \frac{\gamma_j^2}{2} \sum_{\forall k \in \Omega_{m,j}} \left(p^{(k)}_{m-1}(1-p^{(k)}_{m-1})\right)
\end{split}
\end{equation}

Take the derivative of Equation~\ref{eq:taylor-loss} with respect to $\gamma_j$ and set it to zero:

\begin{equation*} 
\begin{split}
    \frac{\partial f(\gamma_j)}{\partial \gamma_{j}} & \approx \sum_{\forall k \in \Omega_{m,j}} \left(-y^{(k)} + p_{m-1}^{(k)}\right) + \gamma_j \sum_{\forall k \in \Omega_{m,j}} 
 \left(p_{m-1}^{(k)}(1-p_{m-1}^{(k)})\right) := 0
\end{split}
\end{equation*}

\begin{equation} \label{eq:gamma-j-complex}
    \Rightarrow \gamma_{j} = \frac{\sum_{\forall k \in \Omega_{m,j}} \left(y^{(k)} - p_{m-1}^{(k)}\right)}{\sum_{\forall k \in \Omega_{m,j}} 
 p_{m-1}^{(k)}\left(1-p_{m-1}^{(k)}\right)} = \frac{\sum_{\forall k \in \Omega_{m,j}} r^{(k)}_m}{\sum_{\forall k \in \Omega_{m,j}} 
 p_{m-1}^{(k)}\left(1-p_{m-1}^{(k)}\right)}
\end{equation}

%% file: contents/summary.tex
\section{Summary}

This document explores the inner workings of the Gradient Boosting Classifier (GBC), a special case of the Gradient Boosting Machine (GBM) specifically designed for binary classification tasks. GBC builds a model by iteratively adding weak learners (decision trees) that predict the residual errors of previous iterations. Unlike GBM used for regression, GBC involves predicting probabilities for binary outcomes and, as such, each terminal node in a tree outputs values determined by the residual errors and prior probabilities from earlier iterations.

We introduce the central equation that determines $\gamma_j$, the output value of a terminal node. The derivation of the terminal node value $\gamma_j$ is explored in detail, demonstrating how it minimizes the logistic loss function. We tried to solve the optimal $\gamma_j$ analytically by setting the derivative of the loss function to zero, though this appeared to be intractable. Consequently, we used a Taylor series approximation to derive a more tractable expression for $\gamma_j$, revealing how it is connected to the residuals and prior predictions of the model.

In conclusion, the document provides a comprehensive overview of the Gradient Boosting Classifier, covering both the theoretical aspects and practical implementation steps while highlighting the role of the terminal node values $\gamma_j$ in optimizing the model's performance in binary classification tasks.

%% file: contents/appendix.tex
\section{Appendix: a working example} \label{app:example}

\begin{table}[tbh]
\centering
\caption{Training instances}
\label{tab:train-instances}
\begin{tabular}{@{}lll@{}}
\toprule
index $i$ & feature $\bm{x}^{(i)}$ & label $y^{(i)}$ \\ \midrule
1       & 1.3        & 1        \\
2       & 1.5        & 0        \\
3       & 3.0        & 1        \\
4       & 4.0        & 0        \\
5       & 6.5        & 1        \\
6       & 8.4        & 0        \\ \bottomrule
\end{tabular}
\end{table}

\begin{table}[tbh]
\centering
\caption{The residuals for iteration 1}
\label{tab:ex-iter1-residual}
\begin{tabular}{@{}rrrrr@{}}
\toprule
index $i$ & $\bm{x}^{(i)}$ & $y^{(i)}$  & $p_0^{(i)}$ & $r_1^{(i)}$ \\ \midrule
1   &1.3  &1  & 0.5 & 0.5       \\
2   &1.5  &0  & 0.5 & $-0.5$    \\
3   &3.0  &1  & 0.5 & 0.5       \\
4   &4.0  &0  & 0.5 & $-0.5$    \\
5   &6.5  &1  & 0.5 & 0.5       \\
6   &8.4  &0  & 0.5 & $-0.5$    \\ \bottomrule
\end{tabular}
\end{table}

We use a simple example to illustrate the training and prediction of GBC. The training dataset is given in Table~\ref{tab:train-instances}. We will train GBC in 3 iterations.

\subsection{Training iteration 0: initialization}

GBC first initializes all $F_0^{(i)}$ as zero, so the predicted $p_0^{(i)}$ are all $\sigma(F_0^{(i)}) = 0.5$ (referring to lines 1 through 3 in Algorithm~\ref{alg:gbc-train}). 

\subsection{Training iteration 1}

Next, we enter iteration 1. GBC computes the residuals $r_1^{(i)}$ for each instance (lines 5 and 6 of Algorithm~\ref{alg:gbc-train}). The result is shown in Table~\ref{tab:ex-iter1-residual}.

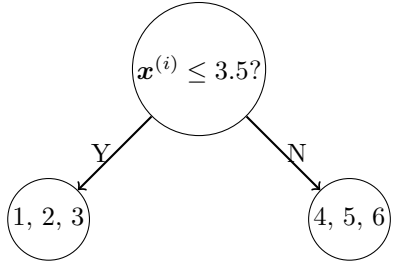
\begin{figure}[tbh]
\centering
\begin{tikzpicture}[
  level distance=2cm, sibling distance=4cm,  
  edge from parent/.style={draw,->,thick},   
  every node/.style={inner sep=1pt},         
  treenode/.style={circle, draw},            
  ]

  \node[treenode] {$\bm{x}^{(i)} \le 3.5$?}
    child {node[treenode] {1, 2, 3} edge from parent node[midway,left] {Y}}
    child {node[treenode] {4, 5, 6} edge from parent node[midway,right] {N}};
\end{tikzpicture}
\caption{Tree $T_1$}
\label{fig:ex-iter1-tree}
\end{figure}

GBC builds a one-level decision tree (a.k.a., a decision stump) $T_1$ to fit the residuals. Assuming the optimal split is at $\bm{x}=3.5$, we have a tree shown in Figure~\ref{fig:ex-iter1-tree}. The numbers inside the terminal nodes indicate the instance index. In the example, instances 1, 2, and 3 are located in the first child of $T_1$ since $\bm{x}^{(i)} \le 3.5$ when $i\in\{1,2,3\}$, and instances 4, 5, 6 are in the second child of $T_1$. We use $\Omega_{m,j}$ to denote the set of instances in the $ T_m$'s $j$th terminal node. Thus, $\Omega_{1,1} = \{1,2,3\}$ and $\Omega_{1,2} = \{4, 5, 6\}$.

For each terminal node $j$, the output value $\gamma_j$ is calculated
by Equation~\ref{eq:gamma-j}. In our example:
\begin{equation}
\begin{split}
    \gamma_1 = \sum_{\forall k \in \Omega_{1,1}} r_1^{(i)} \biggr/ \sum_{\forall k \in \Omega_{1,1}}\left(p_0^{(i)} (1-p_0^{(i)})\right) = 2/3 \\
    \gamma_2 = \sum_{\forall k \in \Omega_{1,2}} r_1^{(i)} \biggr/ \sum_{\forall k \in \Omega_{1,2}}\left(p_0^{(i)} (1-p_0^{(i)})\right) = -2/3 
\end{split}
\end{equation}

The predicted logarithmic odds are updated by Equation~\ref{eq:logloss-update}:

\begin{equation}
\begin{split}
    F_1^{(1)} &= F_0^{(1)} + \eta \gamma_{1,1} = 0+0.1\times \frac{2}{3} = \frac{2}{30} \\
    F_1^{(2)} &= F_0^{(2)} + \eta \gamma_{1,1} = 0+0.1\times \frac{2}{3} = \frac{2}{30} \\
    F_1^{(3)} &= F_0^{(3)} + \eta \gamma_{1,1} = 0+0.1\times \frac{2}{3} = \frac{2}{30} \\
    F_1^{(4)} &= F_0^{(4)} + \eta \gamma_{1,2} = 0+0.1\times \frac{-2}{3} = -\frac{2}{30} \\
    F_1^{(5)} &= F_0^{(5)} + \eta \gamma_{1,2} = 0+0.1\times \frac{-2}{3} = -\frac{2}{30} \\
    F_1^{(6)} &= F_0^{(6)} + \eta \gamma_{1,3} = 0+0.1\times \frac{-2}{3} = -\frac{2}{30} 
\end{split}
\end{equation}

Finally, we update the probabilities.

\begin{equation}
\begin{split}
    p_1^{(1)} &= \sigma\left(F_1^{(1)}\right) = 0.5167\\
    p_1^{(2)} &= \sigma\left(F_1^{(2)}\right) = 0.5167\\
    p_1^{(3)} &= \sigma\left(F_1^{(3)}\right) = 0.5167 \\
    p_1^{(4)} &= \sigma\left(F_1^{(4)}\right) = 0.4833\\
    p_1^{(5)} &= \sigma\left(F_1^{(5)}\right) = 0.4833 \\
    p_1^{(6)} &= \sigma\left(F_1^{(6)}\right) = 0.4833
\end{split}
\end{equation}

\subsection{Training iteration 2}

\begin{table}[tbh]
\centering
\caption{The residuals for iteration 2}
\label{tab:ex-iter2-residual}
\begin{tabular}{@{}rrrrr@{}}
\toprule
index $i$ & $\bm{x}^{(i)}$ & $y^{(i)}$  & $p_1^{(i)}$ & $r_2^{(i)}$ \\ \midrule
1   &1.3  &1  & 0.5167 & 0.4833       \\
2   &1.5  &0  & 0.5167 & $-0.5167$    \\
3   &3.0  &1  & 0.5167 & 0.4833       \\
4   &4.0  &0  & 0.4833 & $-0.4833$    \\
5   &6.5  &1  & 0.4833 & 0.5167       \\
6   &8.4  &0  & 0.4833 & $-0.4833$    \\ \bottomrule
\end{tabular}
\end{table}

We enter iteration 2. GBC computes the residuals $r_2^{(i)} = y^{(i)} - p_1^{(i)}$ for each instance in this iteration. The result is shown in Table~\ref{tab:ex-iter2-residual}.

\begin{figure}[tbh]
\centering
\begin{tikzpicture}[
  level distance=2cm, sibling distance=4cm,  
  edge from parent/.style={draw,->,thick},   
  every node/.style={inner sep=1pt},         
  treenode/.style={circle, draw},            
  ]

  \node[treenode] {$\bm{x}^{(i)} \le 2.25$?}
    child {node[treenode] {1, 2} edge from parent node[midway,left] {Y}}
    child {node[treenode] {3, 4, 5, 6} edge from parent node[midway,right] {N}};
\end{tikzpicture}
\caption{Tree $T_2$}
\label{fig:ex-iter2-tree}
\end{figure}
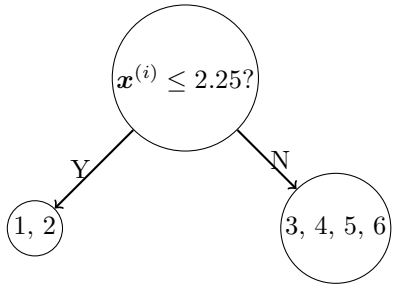

GBC builds another one-level decision tree $T_2$ to fit the residuals. Assuming the optimal split is at $\bm{x}=2.25$, then we have a tree shown in Figure~\ref{fig:ex-iter2-tree}: training instances 1 and 2 are located in the first child of $T_2$, and instances 3, 4, 5, 6 are in the second child of $T_2$, i.e., $\Omega_{2,1} = \{1,2\}$ and $\Omega_{2,2} = \{3, 4, 5, 6\}$.

For each terminal node $j$, the output value $\gamma_j$ is calculated
by Equation~\ref{eq:gamma-j}. So:
\begin{equation}
\begin{split}
    \gamma_1 = \sum_{\forall k \in \Omega_{2,1}} r_2^{(i)} \biggr/ \sum_{\forall k \in \Omega_{2,1}}\left(p_1^{(i)} (1-p_1^{(i)})\right) \approx -0.0669 \\
    \gamma_2 = \sum_{\forall k \in \Omega_{2,2}} r_2^{(i)} \biggr/ \sum_{\forall k \in \Omega_{2,2}}\left(p_1^{(i)} (1-p_1^{(i)})\right) \approx 0.0334
\end{split}
\end{equation}

Next, we can update the model predicted log odds $F_2^{(i)}$ based on Equation~\ref{eq:logloss-update}:

\begin{equation}
\begin{split}
    F_2^{(1)} &= F_1^{(1)} + \eta \gamma_{2,1} = \frac{2}{30}+0.1\times (-0.0669) \approx 0.0600 \\
    F_2^{(2)} &= F_1^{(2)} + \eta \gamma_{2,1} = \frac{2}{30}+0.1\times (-0.0669) \approx 0.0600  \\
    F_2^{(3)} &= F_1^{(3)} + \eta \gamma_{2,2} = \frac{2}{30}+0.1\times 0.0334 \approx 0.0700 \\
    F_2^{(4)} &= F_1^{(4)} + \eta \gamma_{2,2} = -\frac{2}{30}+0.1\times  0.0334 \approx -0.0633 \\
    F_2^{(5)} &= F_1^{(5)} + \eta \gamma_{2,2} = -\frac{2}{30}+0.1\times  0.0334 \approx -0.0633 \\
    F_2^{(6)} &= F_1^{(6)} + \eta \gamma_{2,3} = -\frac{2}{30}+0.1 \times 0.0334 \approx -0.0633
\end{split}
\end{equation}

Finally, we update the probabilities.

\begin{equation}
\begin{split}
    p_2^{(1)} &= \sigma\left(F_2^{(1)}\right) \approx 0.5150 \\
    p_2^{(2)} &= \sigma\left(F_2^{(2)}\right) \approx 0.5150 \\
    p_2^{(3)} &= \sigma\left(F_2^{(3)}\right) \approx 0.5175 \\
    p_2^{(4)} &= \sigma\left(F_2^{(4)}\right) \approx 0.4842 \\
    p_2^{(5)} &= \sigma\left(F_2^{(5)}\right) \approx 0.4842 \\
    p_2^{(6)} &= \sigma\left(F_2^{(6)}\right) \approx 0.4842 
\end{split}
\end{equation}

\subsection{Training iteration 3}

\begin{table}[tbh]
\centering
\caption{The residuals for iteration 3}
\label{tab:ex-iter3-residual}
\begin{tabular}{@{}rrrrr@{}}
\toprule
index $i$ & $\bm{x}^{(i)}$ & $y^{(i)}$  & $p_2^{(i)}$ & $r_3^{(i)}$ \\ \midrule
1   &1.3  &1  & 0.5150 & 0.4850      \\
2   &1.5  &0  & 0.5150 & $-0.5150$   \\
3   &3.0  &1  & 0.5175 & 0.4825      \\
4   &4.0  &0  & 0.4842 & $-0.4842$   \\
5   &6.5  &1  & 0.4842 & 0.5158      \\
6   &8.4  &0  & 0.4842 & $-0.4842$   \\ \bottomrule
\end{tabular}
\end{table}

We enter iteration 3. GBC computes the residuals $r_3^{(i)} = y^{(i)} - p_2^{(i)}$ for each instance in this iteration. The result is shown in Table~\ref{tab:ex-iter3-residual}.

\begin{figure}[tbh]
\centering
\begin{tikzpicture}[
  level distance=2cm, sibling distance=4cm,  
  edge from parent/.style={draw,->,thick},   
  every node/.style={inner sep=1pt},         
  treenode/.style={circle, draw},            
  ]

  \node[treenode] {$\bm{x}^{(i)} \le 5.25$?}
    child {node[treenode] {1, 2, 3, 4} edge from parent node[midway,left] {Y}}
    child {node[treenode] {5, 6} edge from parent node[midway,right] {N}};
\end{tikzpicture}
\caption{Tree $T_3$}
\label{fig:ex-iter3-tree}
\end{figure}
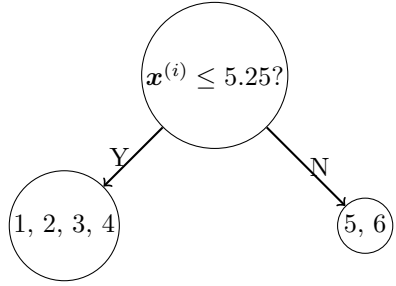

GBC builds yet another decision stump $T_3$ to fit the residuals. Assuming the optimal split is at $\bm{x}=5.25$, then we have a tree shown in Figure~\ref{fig:ex-iter3-tree}: training instances 1, 2, 3, and 4 are located in the first child of $T_3$, and instances 5 and 6 are in the second child of $T_3$, i.e., $\Omega_{3,1} = \{1,2,3,4\}$ and $\Omega_{3,2} = \{5, 6\}$.

For each terminal node $j$, the output value $\gamma_j$ is calculated
by Equation~\ref{eq:gamma-j}. So:
\begin{equation}
\begin{split}
    \gamma_1 = \sum_{\forall k \in \Omega_{3,1}} r_3^{(i)} \biggr/ \sum_{\forall k \in \Omega_{3,1}}\left(p_2^{(i)} (1-p_2^{(i)})\right) \approx -0.0317 \\
    \gamma_2 = \sum_{\forall k \in \Omega_{3,2}} r_3^{(i)} \biggr/ \sum_{\forall k \in \Omega_{3,2}}\left(p_2^{(i)} (1-p_2^{(i)})\right) \approx 0.0633
\end{split}
\end{equation}

Next, we can update the model predicted log odds $F_3^{(i)}$ based on Equation~\ref{eq:logloss-update}:

\begin{equation}
\begin{split}
    F_3^{(1)} &= F_2^{(1)} + \eta \gamma_{3,1} \approx 0.06 + 0.1 \times (-0.0317) \approx 0.0568 \\
    F_3^{(2)} &= F_2^{(2)} + \eta \gamma_{3,1} \approx 0.06 + 0.1 \times (-0.0317) \approx 0.0568 \\
    F_3^{(3)} &= F_2^{(3)} + \eta \gamma_{3,1} \approx 0.07 + 0.1 \times (-0.0317) \approx 0.0668\\
    F_3^{(4)} &= F_2^{(4)} + \eta \gamma_{3,1} \approx (-0.0633) + 0.1 \times (-0.0317) \approx -0.0665 \\
    F_3^{(5)} &= F_2^{(5)} + \eta \gamma_{3,2} \approx (-0.0633) + 0.1 \times 0.0633 \approx -0.0570 \\
    F_3^{(6)} &= F_2^{(6)} + \eta \gamma_{3,3} \approx (-0.0633) + 0.1 \times 0.0633 \approx -0.0570
\end{split}
\end{equation}

Finally, we update the probabilities.

\begin{equation}
\begin{split}
    p_3^{(1)} &= \sigma\left(F_3^{(1)}\right) \approx 0.5142 \\
    p_3^{(2)} &= \sigma\left(F_3^{(2)}\right) \approx 0.5142 \\
    p_3^{(3)} &= \sigma\left(F_3^{(3)}\right) \approx 0.5167 \\
    p_3^{(4)} &= \sigma\left(F_3^{(4)}\right) \approx 0.4834 \\
    p_3^{(5)} &= \sigma\left(F_3^{(5)}\right) \approx 0.4858 \\
    p_3^{(6)} &= \sigma\left(F_3^{(6)}\right) \approx 0.4858 
\end{split}
\end{equation}

\subsection{Prediction}

Eventually, the predicted log odds can be shown in Equation~\ref{eq:pred}, where the terminal nodes of a tree show the predicted log odds of that tree.

\begin{equation} \label{eq:pred}
    \hat{y} = 0 + \eta
    \raisebox{-0.5\height}{
    \begin{tikzpicture}[level distance=1.5cm, sibling distance=2.5cm,
    edge from parent/.style={draw,->,thick},
    every node/.style={inner sep=1pt},
    treenode/.style={circle, draw}]
        \node[treenode] {$\bm{x}^{(i)} \le 3.5$?}
            child {node[treenode] {2/3} edge from parent node[midway,left] {Y}}
            child {node[treenode] {-2/3} edge from parent node[midway,right] {N}};
    \end{tikzpicture}
    }
    + \eta
    \raisebox{-0.5\height}{
    \begin{tikzpicture}[level distance=1.5cm, sibling distance=2.5cm,
    edge from parent/.style={draw,->,thick},
    every node/.style={inner sep=1pt},
    treenode/.style={circle, draw}]
        \node[treenode] {$\bm{x}^{(i)} \le 2.25$?}
            child {node[treenode] {-0.0669} edge from parent node[midway,left] {Y}}
            child {node[treenode] {0.0334} edge from parent node[midway,right] {N}};
    \end{tikzpicture}
    }
    + \eta
    \raisebox{-0.5\height}{
    \begin{tikzpicture}[level distance=1.5cm, sibling distance=2.5cm,
    edge from parent/.style={draw,->,thick},
    every node/.style={inner sep=1pt},
    treenode/.style={circle, draw}]
        \node[treenode] {$\bm{x}^{(i)} \le 5.25$?}
            child {node[treenode] {-0.0317} edge from parent node[midway,left] {Y}}
            child {node[treenode] {0.0633} edge from parent node[midway,right] {N}};
    \end{tikzpicture}
    }
\end{equation}

If given a new test instance with $\bm{x}=7$, the predicted log odds is

\begin{equation}
    \hat{y} = 0 + 0.1 \times \left(-\frac{2}{3}\right) + 0.1 \times 0.0334 + 0.1 \times 0.0633 \approx -0.0570
\end{equation}

The predicted probability that $y=1$ is

\begin{equation}
    p = \sigma\left(-0.0570\right) = 0.4858
\end{equation}

If we set the threshold at $0.5$, the GBC predicts this instance as negative.